\documentclass[letterpaper, 10 pt, conference]{ieeeconf}  

\IEEEoverridecommandlockouts                              

\usepackage[tracking=true]{microtype}
\usepackage{amsmath} 
\usepackage{amssymb}  
\usepackage{graphicx}
\usepackage{epsfig} 
\usepackage{mathptmx} 
\usepackage{times} 
\usepackage{amsmath} 
\usepackage{amssymb}  
\usepackage{xspace}

\usepackage[frozencache,cachedir=minted]{minted}
\setminted{fontsize=\footnotesize}
\usepackage{url}
\usepackage[dvipsnames]{xcolor}
\usepackage[font={footnotesize}]{caption}
\usepackage{booktabs} 
\usepackage{tikz}
\usetikzlibrary{positioning,shapes.geometric,arrows,fit,shapes.symbols,svg.path,tikzmark,calc}
\usepackage{booktabs}
\usepackage{textcomp}
\usepackage{listings}
\usepackage{subcaption}
\pgfdeclarelayer{background}
\pgfdeclarelayer{foreground}
\pgfsetlayers{background,main,foreground}
\usepackage{siunitx}
\usepackage{adjustbox}
\usepackage{array}
\usepackage{multirow}
\usepackage{geometry}

\let\labelindent\relax 

\usepackage[inline]{enumitem} 

\usepackage{xspace}

\newcommand{\todo}[1]{}

\newcommand{\ryan}[1]{}
\newcommand{\caelan}[1]{}
\newcommand{\ajay}[1]{}

\usepackage{algorithm}
\usepackage[noend]{algpseudocode}
\usepackage{amssymb}
\algnewcommand\algorithmicdeclare{\textbf{Declare:}}
\algnewcommand\Declare{\item[\algorithmicdeclare]}

\usepackage{booktabs}
\usepackage{hyperref}
\usepackage{cleveref}
\usepackage{dblfloatfix}
\usepackage{graphicx} 
\usepackage{caption}
\usepackage{pifont}
\usepackage{verbatimbox}
\usepackage{bm}

\usepackage[noadjust]{cite}

\title{ARMADA: Augmented Reality for Robot \\ Manipulation and Robot-Free Data Acquisition \\
}

\author{Nataliya Nechyporenko$^{*12}$, Ryan Hoque$^{*2}$, Christopher Webb$^{2}$, Mouli Sivapurapu$^{2}$, Jian Zhang$^{2}$
\thanks{$^{1}$Human Interaction and Robotics [HIRO] Group at the University of Colorado Boulder. Nataliya was an intern at Apple during this project.} 
\thanks{$^{2}$Apple
}
\thanks{$^*$Equal contribution. Correspondence to {\tt nataliya@colorado.edu, ryanhoque@apple.com}.}}

\begin{document}

\maketitle
\thispagestyle{empty}
\pagestyle{empty}

\begin{abstract}
Teleoperation for robot imitation learning is bottlenecked by hardware availability. Can high-quality robot data be collected without a physical robot? We present a system for augmenting Apple Vision Pro with real-time virtual robot feedback. By providing users with an intuitive understanding of how their actions translate to robot motions, we enable the collection of natural barehanded human data that is compatible with the limitations of physical robot hardware. We conducted a user study with 15 participants demonstrating 3 different tasks each under 3 different feedback conditions and directly replayed the collected trajectories on physical robot hardware. Results suggest live robot feedback dramatically improves the quality of the collected data, suggesting a new avenue for scalable human data collection without access to robot hardware. Videos and more are available at \url{https://nataliya.dev/armada}.  

\end{abstract}

\section{Introduction}

Imitation learning (IL) has become a leading approach for enabling robots to perform complex manipulation tasks \cite{Hussein2017imitation, chi2024diffusion}. In IL, a dataset of human behavior is collected, and the robot is trained to mimic this behavior via supervised machine learning. 
Prior work shows that a robot trained with IL on expert data collected with specialized hardware can autonomously perform dexterous tasks such as inserting batteries, rinsing dishes, and tying shoelaces \cite{zhao2023aloha1, fu2024mobilealoha, zhao2024alohaunleashed}. 

Meanwhile, large language and vision models have recently achieved state-of-the-art results in natural language processing and 2D computer vision via supervised machine learning with Internet-scale datasets \cite{touvron2023llama, kenton2019bert, kirillov2023segment, tu2022maxvit, radford2021learning}. Can a similar approach be applied to robot manipulation? Unfortunately, unlike vision and language, there is no existing Internet-scale dataset for robot control. While some works aim to learn robot control from Internet-scale human video \cite{nair2022r3m, radosavovic2023real, ma2022vip}, these approaches face a prohibitively large embodiment gap between human and robot. Currently, no robot can accurately replicate human hand actions, as the sensitivity, compliance, and actuation of a human hand is superior to any robot end-effector, and the kinematics and dynamics of existing robot arms do not match those of humans. 
As a result, data collection efforts for IL typically involve human teleoperation of physical robots. While effective, this approach is not scalable as it is bottlenecked by the availability of robot hardware platforms. 


New capabilities in augmented reality suggest a compelling possibility: can humans \textit{without} access to a physical robot provide data that is compatible with robots? We propose a system for introducing a digital twin of a robot into augmented reality with Apple Vision Pro. By overlaying a simulated robot on the high-resolution passthrough of Vision Pro, a human demonstrator can collect real-world manipulation data with their bare hands while observing in real time how the downstream robot would behave. This enables rich, two-way feedback between human and robot that conveys critical information such as robot kinematics, dynamics, and speed during trajectory execution. Such feedback may enable human demonstrators to collect data that is more compatible with robot hardware than human hand data collected without this feedback.

\begin{figure}[t!]
  \begin{center}
    \includegraphics[width=0.99\linewidth]{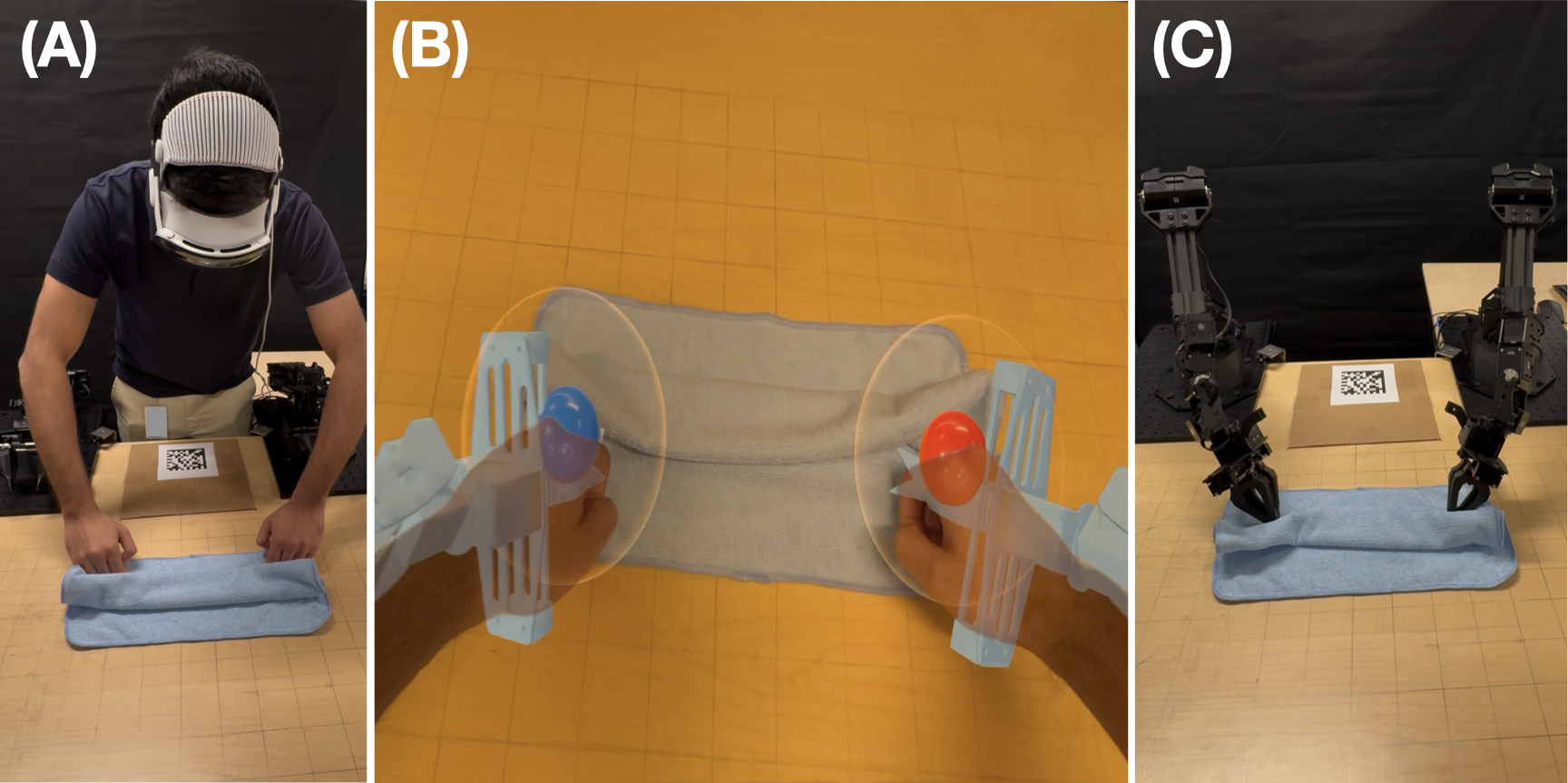}
  \end{center}
  \caption{\textbf{Overview.} (A) Human demonstrators wearing Apple Vision Pro can collect data directly with their hands. (B) Egocentric view within Vision Pro shows real-time robot execution overlaid on the user's hands with augmented reality. (C) High-quality demonstrations collected with this system can be directly replayed on physical robot hardware.}
  \label{fig:teaser}
  \vspace{-15pt}
\end{figure}

To enable new users to interact with the system, we develop ARMADA, a user-friendly software application\footnote[3]{Code will be available on the project website when finalized.} that can be used by anyone with access to an Apple Vision Pro to collect data.  
In a user study with 15 participants, including participants with no prior experience using a virtual reality device, we collect a total of 675 robot-free demonstrations on 3 tasks with our system. Results suggest that demonstrations collected with real-time AR feedback can be directly replayed on physical robot hardware, dramatically increasing the average replay success rate from 1.3\% to 71.1\% when compared to demonstrations collected without such feedback.

By enabling in-the-wild data collection from anyone with a Vision Pro, ARMADA may facilitate the creation of large datasets with tens of thousands of hours of manipulation data, an order of magnitude larger than existing datasets collected with robot teleoperation \cite{padalkar2023open}. Such datasets may enable imitation learning at unprecedented scale, an essential ingredient for generalization across tasks, environments, and robot hardware \cite{padalkar2023open, octo_2023, kim2024openvla}.

\section{Related Work}

\subsection{Data collection for imitation learning}
Although robots can be engineered to perform tasks in structured environments, general-purpose dexterous robot manipulation that adapts to substantial variations in task or environment remains challenging \cite{cui2021toward}. Dexterity is algorithmically challenging due to the high-dimensional action spaces and complex contact dynamics involved \cite{huang2021efficient, cheng2024enhancing}. In robot learning, the challenges from contact modeling and analysis are implicit, but require a careful design of the model architecture and the data collection process. In particular, imitation learning relies on the collection of expert demonstrations to guide the robot's learning process. A key challenge for imitation learning is scaling up the training data.

To help address this challenge, teleoperation has emerged as a promising paradigm. Teleoperation allows the collection of expert demonstrations by enabling operators to remotely control a robot or virtual agent while performing tasks. Various teleoperation approaches have been studied using gloves \cite{wang2024dexcap, liu2017glove, liu2019high}, cameras \cite{handa2020dexpilot, qin2023anyteleop, sivakumar2022robotic, li2020mobile, lin2024learning}, motion capture \cite{zhao2012combining, liu2021semi}, hardware twins, \cite{zhao2023aloha1, fu2024mobilealoha, zhao2024alohaunleashed, wu2023gello}, and VR devices \cite{cheng2024open, ding2024bunny}. However, teleoperation requires the operators and the robot to be connected in a low-latency control loop. Although works like Open-TeleVision \cite{cheng2024open} enable remote teleoperation, they suffer from the limitations of the one-to-one hardware-to-demonstrator requirement. In contrast, our work requires only a kinematic simulation and Apple Vision Pro, a personal device that already exists in thousands of homes. 
 
Other works attempt to learn directly from human video demonstrations, removing the need for robot data collection altogether \cite{duan2023ar2}. 
\todo{we might have to say more about ar2-d2 here since its highly relevant} 
While passive human videos can provide insight into how a human would accomplish a desired task \cite{bharadhwaj2024towards, xu2023xskill, wang2023mimicplay, yang2022learning, xiong2023robotube}, their utility is bottlenecked the ability to `translate' the human action into the robot's embodiment.
Data collected in the camera space with highly articulated human hands \cite{yang2022learning, Ego4D2022CVPR, damen2020epic} often needs to be paired with a sequence of teleoperation demonstrations to bridge the embodiment gap \cite{xu2023xskill, wang2023mimicplay, bharadhwaj2024towards}. 

\subsection{Robot manipulation via augmented or virtual reality}

Recent surveys have highlighted the growing importance of using virtual reality (VR) and augmented reality (AR) for robot manipulation \cite{fu2023recent,su2023integrating}. Several works have demonstrated the utility of AR/VR interfaces for controlling robots, particularly in hazardous conditions. These interfaces often involve using virtual markers or grippers to control the robot's motion in end-effector space \cite{jang2021virtual,van2024puppeteer}. 
While showing a previews of robot motions in AR/VR before execution can reduce cognitive load \cite{meng2023virtual}, the headset may result in increased physical strain after continual use \cite{audonnet2024immertwin,audonnet2024telesim}. 
VR/AR interfaces have also been compared with traditional input devices such as space mice and kinesthetic teaching \cite{smith2024augmented}.
Although these works demonstrate the usefulness of AR/VR for robot manipulation, they primarily focus on virtual teach pendants. In contrast, our work introduces a real-time digital twin of a robot system in AR. 

AR2-D2 \cite{duan2023ar2} pioneered an iOS app for providing robot demonstrations without a physical robot, but it does not provide real-time robot feedback or the intuitive egocentric view of Vision Pro. In concurrent work, Park et al. introduce DART \cite{park2024dexhubdartinternetscale}, an augmented reality data collection system. However, DART is designed for data collection in simulation, which introduces a sim-to-real gap when deployed to physical environments. Similar to \cite{yang2024arcade}, we consider scalability through AR, but we focus on the collection of expert hand demonstrations rather than data expansion. The modeling of hand-object contact is crucial to the success of IL and cannot be expanded by their system.

Most similar to our work is ARCap \cite{chen2024arcap}, another portable data collection system with real-time AR feedback developed concurrently with ours. However, the ARCap system consists of several unwieldy hardware components including a VR headset, an additional RGB-D camera mounted on the headset, motion capture gloves, and VR controllers mounted on the wrists. In contrast, we require only an Apple Vision Pro, enabling barehanded data collection unobstructed by hand or wrist mounts that may be easier to collect as well as more amenable to the demonstration of to a wider range of tasks.





\section{System Design}\label{sec:methods}

\begin{figure*}[t]
\centerline{\includegraphics[width=\textwidth]{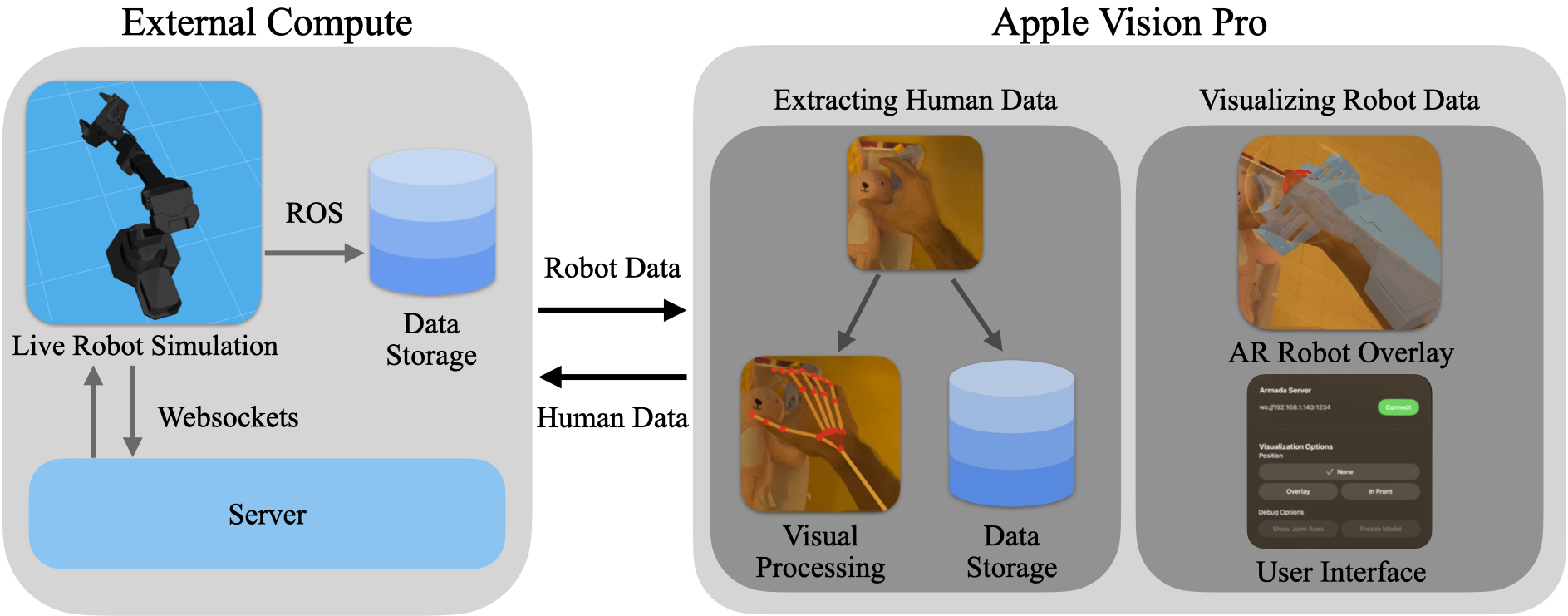}}
\caption{Overview of the system architecture described in Section~\ref{ssec:arch}. Human skeletal data is sent over websockets to an external compute device, which runs a live robot simulation that follows the pose targets given by the human data. The robot proprioceptive data is then sent back to Vision Pro for AR visualization. The full loop runs at 30 Hz.}
\label{fig:system}
\vspace{-10pt}
\end{figure*}


\subsection{Software system architecture}\label{ssec:arch}
The ARMADA app on Apple Vision Pro and the robot control node (on an external compute device) communicate with each other using a combination of ROS and websockets, as shown in \cref{fig:system}. This communication system enables a plug-and-play interface for different hardware platforms. For example, the system can integrate a Franka or UR5 robot by simply swapping the 3D model files and piping the robot information in the expected format to the headset device. In addition, the channels provide a rich set of tools to expand the information being passed and visualized. 

The ARMADA app serves three important purposes: visualization of robots and constraints, data storage, and a user interface. First, to visualize the robot, the ARMADA app receive the transformation frames of every robot link as well as the robot's constraints from the external compute. The base of the robot is positioned relative to a QR code on the table, and each subsequent link is added relative to the base frame. The visualization of constraints is further described in \cref{sec:aug-feedback}. Second, the app captures image frames from the front-facing cameras as well as the human skeleton, which can be exported for data processing. Finally, the app provides a user interface feature to control the visualization and data collection pipeline (Figure~\ref{fig:ui}). 

The robot node also serves three purposes: robot control, data storage, and constraint calculation. The information from the human skeleton is transformed into robot control commands, as described in \cref{sec:robot-arm-contro}. Proprioceptive data (joint angles, velocity, and torque) as well as any external camera frames are stored locally on the external compute. Finally, as the robot is executing an action, the information about any singularity, speed, or workspace violations are calculated in the control loop to be passed to the ARMADA app for visualization.

\subsection{Robot arm control} \label{sec:robot-arm-contro}
The robot is controlled to move according to the human's finger and wrist positions, tracked via ARKit in visionOS 2.0. We found that using the detected wrist position provides a valid position of the human's intended manipulation target, but lacks the valid orientation since the point falls just below the wrist bend. Therefore, we use the average of the four knuckle positions to track the human's position and orientation. The position is then shifted slightly towards the thumb and offset by angle to place the human target directly at the center between the thumb and the index. We use Drake \cite{drake} to compute inverse kinematics (IK) to solve for the joint positions given
the desired end-effector pose (i.e., the human hand pose), and send the joint position command to the robot arm. In the case of IK solving failure, we use the last commanded joint position.

The distance between the index finger and the thumb is used to control the opening and closing of the gripper. If the distance between the finger falls below a certain threshold, the gripper is closed; if the distance increases beyond this threshold, the gripper is opened. 

Together, these two arm control systems simulate the robot hand as a two finger pincher robot. The robot execution loop runs at a rate of 30 Hz.

\subsection{Augmented reality feedback methods} \label{sec:aug-feedback}
\begin{figure}[b]
\centerline{\includegraphics[width=0.5\textwidth]{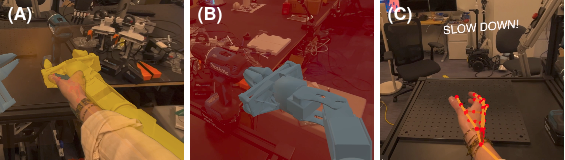}}
\caption{\textbf{Constraints.} (A) The virtual robot gradually turns yellow as it approaches a singular configuration. (B) The background turns red when the robot moves beyond the Cartesian space boundaries. (C) The user is alerted with a virtual text overlay when their hand motion exceeds the robot's velocity limits. }
\label{fig:constraints}
\end{figure}

To demonstrate the system's flexible architecture, we extend its capabilities by enabling the robot to convey additional information to the user. \cref{fig:constraints} illustrates the user's visual field when the robot triggers specific constraints. The following section discusses the significance of this visualization and its potential applications.

\textbf{Singularity} A singularity occurs when a robot loses one or more degrees of freedom (DOF) at a specific point in its workspace. Singularities cause the robot to lose the ability to move freely in certain directions, leading to control issues, precision loss, and potential damage. To check for singularities, we compute the determinant of $J^TJ$, where $J$ is the manipulator Jacobian. When the resulting value falls below a certain threshold, the virtual robot starts to gradually transition into a yellow color, smoothly indicating the proximity to a singular configuration.

\textbf{Workspace Constraints}
Robot manipulators and their vision systems operate within a defined workspace range. If the demonstrator's hands leave this space, tracking may be lost, resulting in inaccurate data. To visualize this constraint violation, a red ``wall" appears at the workspace bounds' maximum or minimum position. 

\textbf{Speed}
To ensure accurate camera tracking when collecting data, the demonstrator needs to move slowly and consistently. Rapid motion can cause blur, which may mean having to discard the collected data sample. If the robot moves too quickly, a ``Slow Down" text appears at the user's eye level. 

While all of these additional feedback modalities are implemented in ARMADA, the user study experiments focus on the subset of modalities described in Section~\ref{ssec:feedback} to limit the fatigue and time commitment of the user study participants.

\begin{figure*}[t]
\centerline{\includegraphics[width=\textwidth]{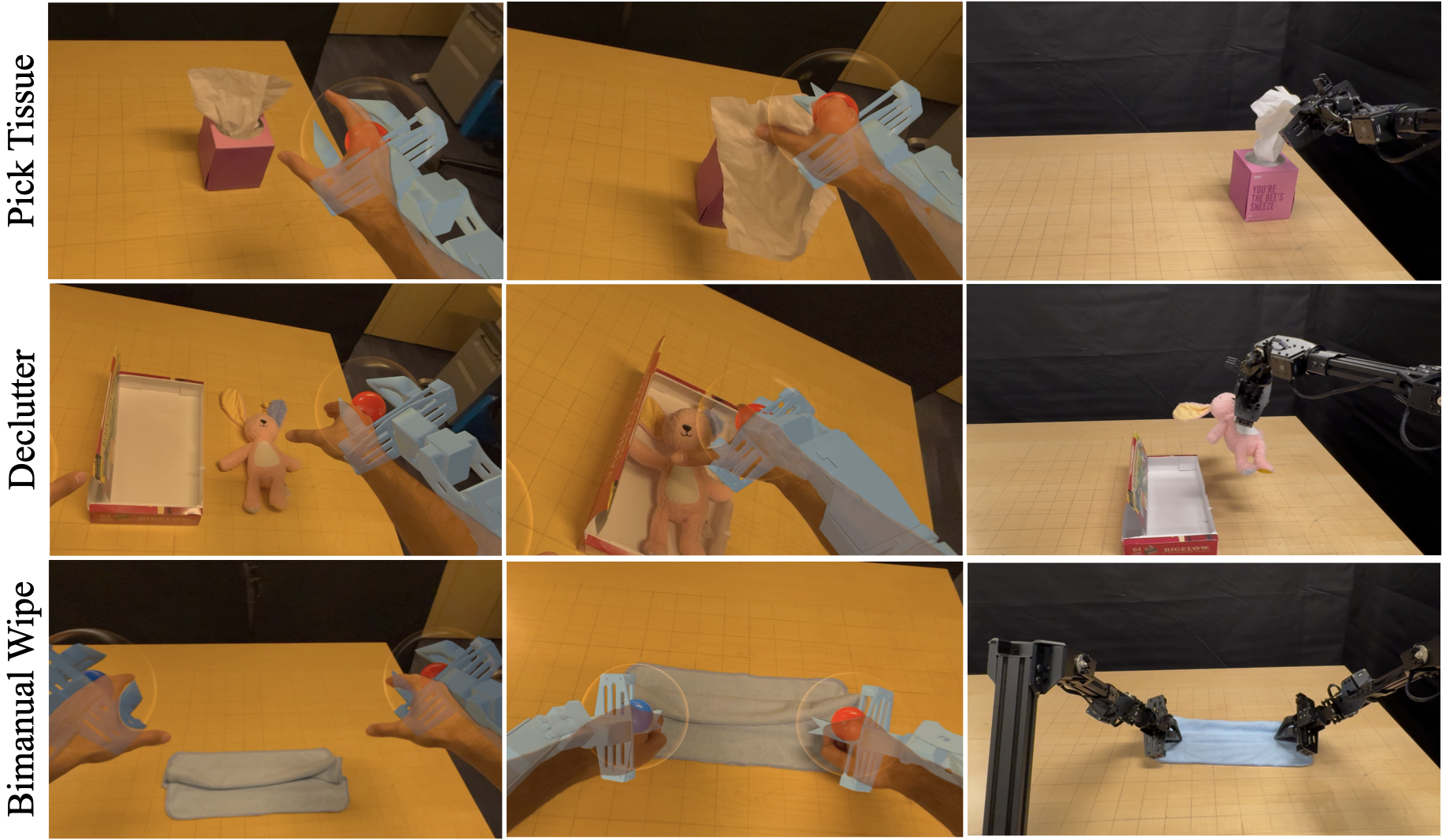}}
\caption{\textbf{Tasks.} The three tasks from Section~\ref{ssec:tasks}. \textit{Left:} egocentric view of initial states during data collection with Feedback. \textit{Middle:} egocentric view of final states during data collection with Feedback. \textit{Right:} mid-task robot execution during trajectory replay.}
\label{fig:tasks}
\vspace{-10pt}
\end{figure*}
\section{User Study}\label{sec:study}

We recruited 15 participants, 11 male and 4 female, with age ranging from 25 to 64. Participants have varying levels of experience with virtual reality devices and robotics, from none at all to significant experience (exact breakdown in Section~\ref{sec:exps}). We describe the experimental setup for the user study below.

\subsection{Feedback Types}\label{ssec:feedback}

The participants provide demonstrations with their hands under three feedback conditions in the following order:

\begin{enumerate}
    \item \textbf{No Feedback:} The user does not see any AR feedback of the robot. The user is instructed to demonstrate the task with natural human motion.
    \item \textbf{Feedback:} The user sees an AR digital twin of the robot updated in real time. The user is instructed to demonstrate the task such that the virtual robot is controlled to execute the task. 
    \item \textbf{Post Feedback:} The user once again cannot see any AR feedback of the robot. The user is once again instructed to demonstrate the task such that the virtual robot is controlled to execute the task. However, since the robot is no longer visible, the user must make an educated guess based on their experience with the Feedback condition as to how the robot is responding to their motions. 
\end{enumerate}

\subsection{Tasks}\label{ssec:tasks}

The participants provide demonstrations for the following three tasks. After data collection, the robot trajectories corresponding to the human arm movements are replayed directly on the robot hardware. See Figure~\ref{fig:tasks} for images of the tasks.

\begin{enumerate}
    \item \textbf{Pick Tissue:} With one hand, grasp a tissue and pull it out of a tissue box. Robot execution is deemed successful if the tissue has been removed from the box.
    \item \textbf{Declutter:} With one hand, pick a soft toy from the tabletop and place it into a cardboard box. Robot execution is deemed successful if the toy has been placed into the cardboard box.
    \item \textbf{Bimanual Wipe:} With two hands, wipe a cloth across the table surface in the direction away from the demonstrator. Robot execution is deemed successful if both end effectors push the cloth away from the respective bases of the robot arms.
    
\end{enumerate}

Each task has 5 distinct starting states that consist of a unique tissue box position, toy position, and cloth position respectively.

\subsection{Experiment Protocol}

\begin{figure}[ht!]
\centerline{\includegraphics[width=\linewidth]{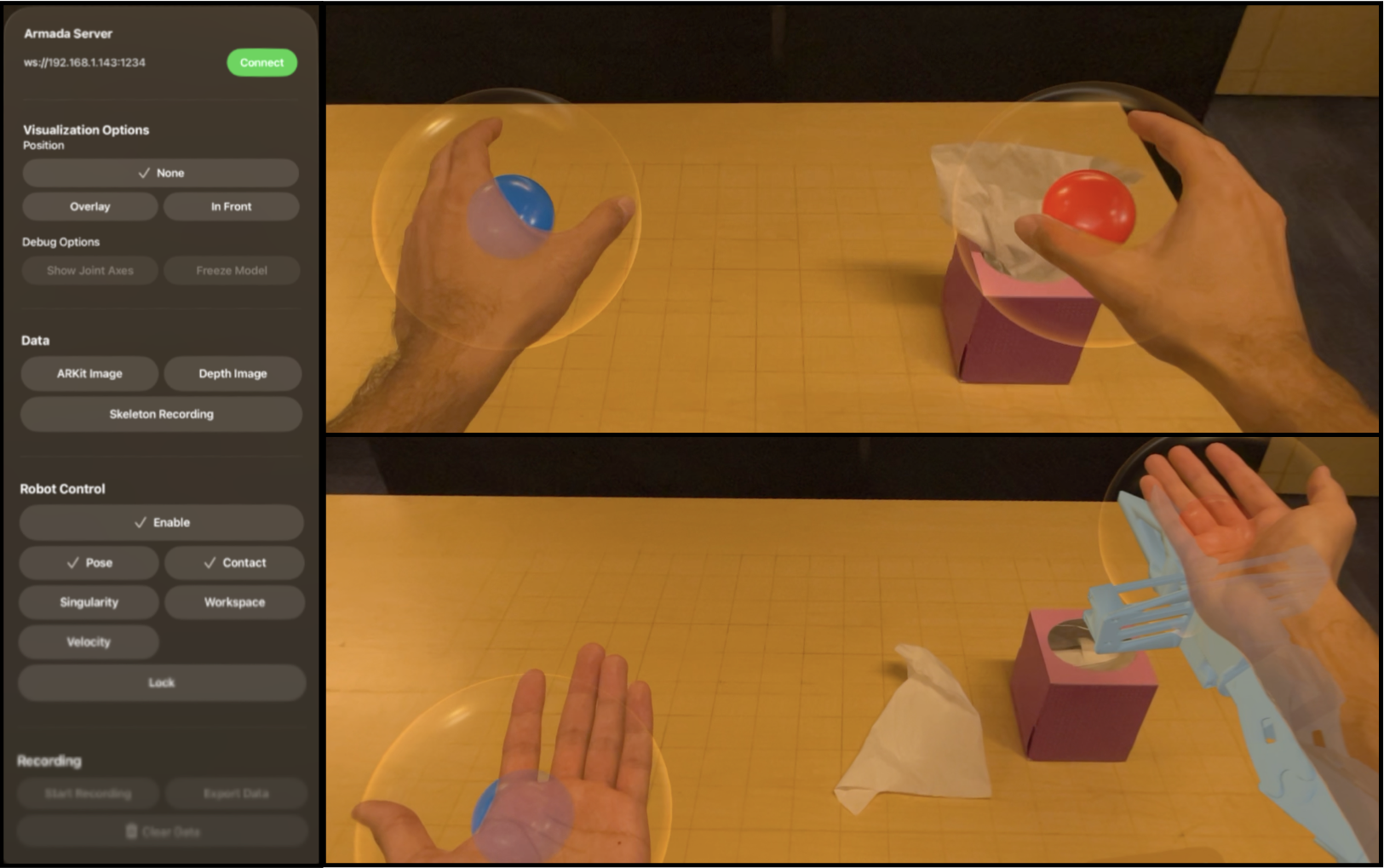}}
\caption{\textbf{User Interface.} \textit{Left:} main menu with toggle options. \textit{Right, Top:} hand placement in spheres to initialize demonstration. \textit{Right, Bottom:} hand placement to release spheres and end demonstration.}
\label{fig:ui}
\vspace{-10pt}
\end{figure}

Each user wears an Apple Vision Pro running the ARMADA software application on visionOS 2.0. The user provides demonstrations under each of the three feedback conditions in order (No Feedback, Feedback, and Post Feedback). During each feedback condition, the user provides demonstrations for all three tasks. The ordering of the tasks is randomly determined for each user and then held fixed for each condition. Since each task has 5 initial states, the user provides 45 demonstrations in total: 5 demonstrations for each of the 3 tasks under No Feedback, followed by 5 $\times$ 3 under Feedback, followed by 5 $\times$ 3 under Post Feedback. After No Feedback and prior to Feedback, the user is given a trial period of 1 minute to get accustomed to the Feedback condition. To begin a demonstration, the user places their hands within two AR spheres in front of them, which then attach to their hands as they demonstrate. To end a demonstration, the user holds their palms face up for 3 seconds, after which the spheres detach from the user's hands and return to their original locations. See Figure~\ref{fig:ui} for images of the user interface. 

During each demonstration, as described in Section~\ref{sec:methods}, wrist and finger poses are estimated in real time with ARKit and streamed to a server. Robots are controlled in simulation to follow these pose targets with IK. The robot states are streamed back to the Vision Pro; under the Feedback condition, the robots are visible. The joint angles of each trajectory are recorded for trajectory replay. At the end of the experiment, the user fills out a short survey about their experience. The full data collection session takes about 45 minutes per user. Afterwards, for evaluation, all 45 saved trajectories are executed on the robot hardware with the environment reset to the appropriate starting states.

\subsection{Robot System}

The robot system consists of a pair of 6 DOF Interbotix ViperX 300 S robot arms, each equipped with a 1 DOF pinch gripper. During data collection, trajectories are executed in simulation. During trajectory replay, the robot trajectories are executed on physical hardware. For the Pick Tissue and Declutter task, only the right arm is actuated. For the Bimanual Wipe task, both arms are actuated. See Figure \ref{fig:teaser}(C) for the hardware setup.
\section{Experiments}\label{sec:exps}

\subsection{Results}

\begin{table}[t!]
    \centering
    \resizebox{\columnwidth}{!}{
    \begin{tabular}{l | r r r}
        \textbf{Feedback Type} & \textbf{Pick Tissue} & \textbf{Declutter} & \textbf{Bimanual Wipe}
        \\ \hline 
        No Feedback & $2.7\% \pm 6.8\%$ & $0.0\% \pm 0.0\%$ & $1.3\% \pm 5.0\%$ \\
        Feedback & $\mathbf{78.7\% \pm 17.1\%}$ & $\mathbf{48.0\% \pm 21.7\%}$ & $\mathbf{86.7\% \pm 15.8\%}$ \\
        Post Feedback & $37.3\% \pm 27.2\%$ & $16.0\% \pm 19.6\%$ & $46.7\% \pm 23.9\%$ \\
    \end{tabular}}
    \caption{\textbf{Replay Success Rate:} Direct trajectory replay success rates for each combination of feedback type and task. Success rate for each participant is computed out of 5 trials, and the mean and standard deviation are computed across 15 user study participants.}
    \label{tab:main}
    \vspace{-5pt}
\end{table}

Here we summarize the key takeaways from Table~\ref{tab:main} and survey results. 



\textbf{Demonstrations collected \textit{with} feedback visualization attain a replay success rate significantly higher than those collected \textit{without} feedback visualization.} In Table~\ref{tab:main}, we observe that Feedback dramatically improves the replay success rate over No Feedback by 76\%, 48\%, and 85\% for the Pick Tissue, Declutter, and Bimanual Wipe tasks, respectively. Demonstrations collected without any feedback have zero or near-zero replayability, suggesting visual robot feedback is critical for collecting high-quality human demonstrations without robot teleoperation.

\textbf{Demonstrations collected \textit{after} feedback visualization attain a higher replay success rate over those collected \textit{without} feedback visualization.} In Table~\ref{tab:main}, we observe that the Post Feedback condition significantly improves upon the replay success rate of No Feedback by 35\%, 16\%, and 45\% for the Pick Tissue, Declutter, and Bimanual Wipe tasks, respectively. This suggests that human demonstrators can \textit{learn} to improve their demonstrations after sufficient experience with live robot feedback, even when these visual cues are subsequently removed. This may further facilitate scalability by lifting the requirement of a live virtual robot in the loop, provided that the demonstrators get sufficient practice with one beforehand.

\textbf{Demonstrations collected \textit{with} feedback visualization attain a higher replay success rate over those collected \textit{after} feedback visualization.} In Table~\ref{tab:main}, we observe that the Feedback condition outperforms the Post Feedback condition by 41\%, 32\%, and 40\% for the Pick Tissue, Declutter, and Bimanual Wipe tasks, respectively. Although Post Feedback improves upon No Feedback, it remains difficult to demonstrate accurately without real-time visual feedback. Results suggest that live robot visualization is still critical for high degrees of replayability.

\textbf{Robot feedback visualization is intuitive and easy to understand.} In survey results, study participants indicated their level of agreement with the statement ``The robot visualization was intuitive and easy to understand." Responses were recorded with a 7-point Likert scale, with 1 corresponding to Completely Disagree and 7 corresponding to Completely Agree. Mean and standard deviation response across all participants was $6.4 \pm 0.6$, indicating a high level of agreement with the statement.

\textbf{Robot feedback visualization is useful for understanding robot motion.} In survey results, study participants indicated their level of agreement with the statement ``The robot visualization was useful for understanding robot motion. I would not be able to predict the motions of the robots without it." Responses were recorded with a 7-point Likert scale, with 1 corresponding to Completely Disagree and 7 corresponding to Completely Agree. Mean and standard deviation response across all participants was $6.0 \pm 1.2$, indicating a high level of agreement with the statement.

\textbf{The system does not require prior experience with robotics or virtual reality devices.} In survey results, participants indicated their level of experience with robotics and prior use of any virtual reality or augmented reality devices. For robotics experience, 2/15 selected None, 3/15 selected Beginner, 3/15 selected Intermediate, 3/15 selected Advanced, and 4/15 selected Expert. For the total duration of the prior use of any VR or AR device, 1/15 had absolutely none, 3/15 had less than 1 hour, 6/15 had between 1 and 5 hours, 
1/15 had between 5 and 20 hours, and 4/15 had more than 20 hours. Results indicate that the system accommodates a wide range of experience levels ranging from none to expert.


\subsection{Analysis}

\textbf{Failure Modes.} Direct replay failure modes for demonstrations collected \textit{without} or \textit{after} robot feedback include excessive speed, imprecise robot pose, undetected opening or closing of the gripper (e.g., due to hand occlusion), and unpredictable inverse kinematics. Replay failures of demonstrations collected \textit{with} robot feedback largely consist of imprecise robot positions or orientations (e.g., due to inaccurate depth perception of the AR display), as even slight imprecision at any point of the trajectory can result in unsuccessful direct replay. For instance, the Declutter task has consistently lower success rates as it requires higher precision than the other tasks. 

\textbf{Qualitative Results.} In survey results, study participants were asked to self-evaluate in freeform text how they behaved differently when collecting demonstrations with robot feedback as opposed to without it. Participants reported slowing down, adjusting their hand positions and orientations, accounting for the opening and closing of the gripper, changing approach angles, and focusing on the robot end effectors rather than their own hands. When asked about the After Feedback condition, participants indicated that they felt they were able to mentally visualize the AR feedback after it had been removed.


\section{Conclusion}

We present a system for high-quality data collection with Apple Vision Pro via real-time AR feedback of robot execution. Results suggest such feedback is critical for collecting robot-free demonstrations that are compatible with real robot kinematics, dynamics, and control. 

One limitation of our work is that evaluation is limited to relatively short-horizon manipulation tasks with compliant objects. More complex tasks may be performed with more experienced demonstrators and improved retargeting and inverse kinematics. 

An important direction for future work is the training of robot control policies on this data via imitation learning. Since human hand poses are estimated by Vision Pro and the pose of the virtual robot is known, the egocentric visual observations collected on Vision Pro can be transformed into robot observations via precise masking and inpainting. Recent work demonstrates that such a technique enables effective policy learning \cite{kareer2024egomimicscalingimitationlearning, wang2024dexcap}.

\bibliographystyle{IEEEtran}
\bibliography{IEEEabrv,references}

\clearpage

\end{document}